\documentclass[letterpaper]{article} 
\usepackage{aaai24}  
\usepackage{times}  
\usepackage{helvet}  
\usepackage{courier}  
\usepackage[hyphens]{url}  
\usepackage{graphicx} 
\usepackage{natbib}  
\usepackage{caption} 
\usepackage{algorithm}
\usepackage{algorithmic}
\usepackage{multirow}
\usepackage{amsmath}
\usepackage{amsfonts} 
\usepackage{mathrsfs}
\usepackage{newfloat}
\usepackage{listings}
\DeclareCaptionStyle{ruled}{labelfont=normalfont,labelsep=colon,strut=off} 
\floatstyle{ruled}
\newfloat{listing}{tb}{lst}{}
\floatname{listing}{Listing}
\title{DiffusionEdge: Diffusion Probabilistic Model for Crisp Edge Detection}

\author{
	Yunfan Ye\textsuperscript{\rm 1,2}\equalcontrib,
	Kai Xu\textsuperscript{\rm 2}\equalcontrib,
	Yuhang Huang\textsuperscript{\rm 2}\thanks{Corresponding author.},
	Renjiao Yi\textsuperscript{\rm 2},
	Zhiping Cai\textsuperscript{\rm 2}
}
\affiliations{
	\textsuperscript{\rm 1}Hunan University\\
	\textsuperscript{\rm 2}National University of Defense Technology
}

\usepackage{bibentry}

\begin{document}

\maketitle

\begin{abstract}
	Limited by the encoder-decoder architecture, learning-based edge detectors usually have difficulty predicting edge maps that satisfy both correctness and crispness. 
	With the recent success of the diffusion probabilistic model (DPM), we found it is especially suitable for accurate and crisp edge detection since the denoising process is directly applied to the original image size. Therefore, we propose the first diffusion model for the task of general edge detection, which we call DiffusionEdge. 
	To avoid expensive computational resources while retaining the final performance, we apply DPM in the latent space and enable the classic cross-entropy loss which is uncertainty-aware in pixel level to directly optimize the parameters in latent space in a distillation manner. 
	We also adopt a decoupled architecture to speed up the denoising process and propose a corresponding adaptive Fourier filter to adjust the latent features of specific frequencies.
	With all the technical designs, DiffusionEdge can be stably trained with limited resources, predicting crisp and accurate edge maps with much fewer augmentation strategies. Extensive experiments on four edge detection benchmarks demonstrate the superiority of DiffusionEdge both in correctness and crispness. On the NYUDv2 dataset, compared to the second best, we increase the ODS, OIS (without post-processing) and AC by 30.2\%, 28.1\% and 65.1\%, respectively. Code: https://github.com/GuHuangAI/DiffusionEdge.
\end{abstract}

\section{Introduction}
Edge detection is a longstanding vision task for detecting object boundaries and visually salient edges from images. As a fundamental problem, it benefits various downstream tasks ranging from 2D perception~\cite{zitnick2014edge, revaud2015epicflow, cheng2020boundary}, generation~\cite{nazeri2019edgeconnect, xiong2019foreground}, and 3D curve reconstruction~\cite{ye2023nef}.

There are three main challenges in general edge detection, \textit{correctness} (identifying edge and non-edge pixels on noisy scenes), \textit{crispness} (the width of edge lines, precisely localizing edges without confusing pixels) and \textit{efficiency} (the inference speed). Traditional methods extract edges based on local features such as gradient~\cite{kittler1983accuracy, canny1986computational}, which can be crisp but not correct enough. Deep learning-based methods~\cite{xie2015holistically, liu2017richer, he2019bi, poma2020dense, pu2022edter, zhou2023treasure} achieve significant progress by capturing local and global features with multi-layers, which is correct but not crisp enough. 
Recently, efforts have also been made to design lightweight architectures~\cite{su2021pixel} for efficiency, or loss functions~\cite{deng2018learning, huan2021unmixing} and refinement strategies~\cite{ye2023delving} for crisp edge detection. However, none of each single edge detector can directly predict edge maps that simultaneously satisfy both correctness and crispness, without a post-processing of morphological non-maximal suppression (NMS) scheme. We
ask this question: Can we learn an edge detector that can directly generate both accurate and crisp edge maps without heavily relying on post-processing?

\begin{figure}[t]
	\centering
	\includegraphics[width=1.0\columnwidth]{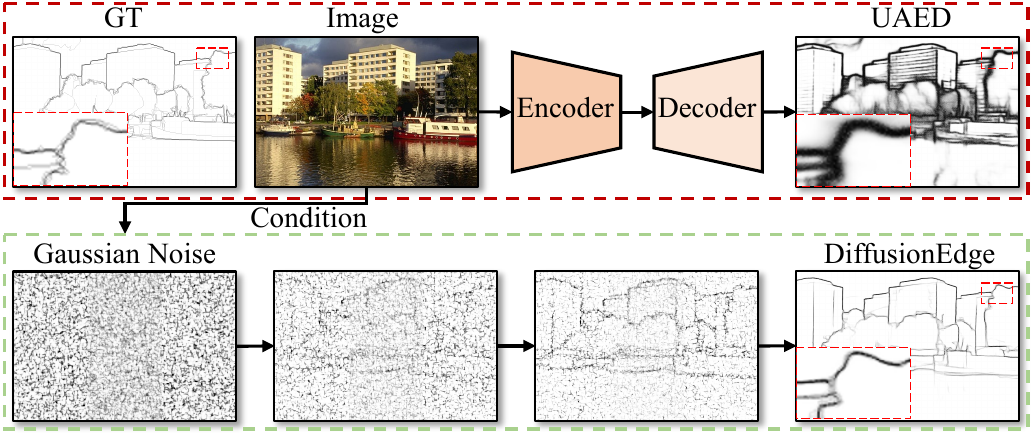} 
	\caption{CNN-based methods, even the most recent and state-of-the-art one (UAED~\cite{zhou2023treasure}), generally have an encoder-decoder architecture with limitations of thick edges and more noise. We propose the diffusion-based edge detector which is superior in both correctness and crispness without any post-processing.}
	\label{fig:teaser}
	\vspace{-0.56cm}
\end{figure}

In this work, we try to answer the question through learning a diffusion model for edge detection. As demonstrated in Figure~\ref{fig:teaser}, DPMs have two main differences compared with methods based on the Convolutional Neural Network (CNN): (a) CNN-based models generally learn and infer the targets in a single round, while DPMs are trained to predict a denoised variant of the noisy input by several steps, which makes it easier for DPMs to learn the target distribution; 
(b) CNN-based edge detectors generally extract features from multi-layers and therefore are limited by the existence of downsampling (for high-level global features) and upsampling (for pixel-wise alignment) operators, which leads to thick edge predictions in nature~\cite{huan2021unmixing}, while DPMs directly perform the denoising process on the level of original image size. 

With the two characteristics, we found diffusion model is especially suitable for accurate and crisp edge detection. However, there are still several challenges for DiffusionEdge to be accurate and crisp enough with limited computational resources and inference time. 
We apply a decoupled diffusion architecture similar to DDM~\cite{huang2023decoupled} to speed up the inference, and propose an adaptive Fourier filter before decoupling, which enables the network weights to adjust the components of the specific frequencies adaptively. 
Following~\cite{rombach2022high}, we also train the diffusion model in latent space to reduce computations. However, most CNN-based edge detectors are trained by the annotator-robust cross entropy loss~\cite{liu2017richer} in image pixel level, which provides uncertainty information when training edge datasets labeled by several annotators like BSDS~\cite{arbelaez2010contour}. To keep that free and valuable uncertainty prior, we apply an uncertainty distillation strategy by directly passing the optimized gradients from pixel level to latent space level based on the chain rule.  

With the above efforts, extensive experiments on four edge detection benchmarks show that DiffusionEdge can directly generate accurate and crisp edge maps without any post-processing, and achieve superior qualitative and quantitative performance with much less augmentation strategies. On the NYUDv2 dataset~\cite{silberman2012indoor}, compared to the second best, we increase the ODS, OIS (without post-processing) and AC by 30.2\%, 28.1\% and 65.1\%, respectively. Our contributions include:
\begin{itemize}
	\item A novel diffusion-based edge detector, named DiffusionEdge, which can predict accurate and crisp edge maps without post-processing. To our best knowledge, it is the first diffusion model toward edge detection.
	\item Several technical designs to ensure learning a satisfactory diffusion model in latent space, while keeping the uncertainty prior and adaptively filtering latent features in Fourier space.
	\item Superior performance on four edge detection benchmarks for both correctness and crispness.
	
\end{itemize}
\section{Related Work}
\paragraph{Edge detection.} 
Edge detection aims to extract object boundaries and visually salient edges from natural images. Traditional edge detectors as such Sobel \cite{kittler1983accuracy} and Canny \cite{canny1986computational} generate edges through local gradients, which often suffer from noisy pixels without global content. 
CNN-based methods start integrating features from multi-layers and improve the correctness of edge pixels by a large margin. HED~\cite{xie2015holistically} proposed the first end-to-end edge detection architecture, and RCF~\cite{liu2017richer} improved it by integrating more hierarchical features. BDCN~\cite{he2019bi} trains the edge detector with layer-specific supervisions in a  bi-directional cascade architecture. PiDiNet~\cite{su2021pixel} introduced pixel difference convolution in the designed lightweight architectures for efficient edge detection. UAED~\cite{zhou2023treasure} measures the degree of
ambiguity among different annotations from multiple annotations to focus more on hard samples.  Also, EDTER~\cite{pu2022edter} proposed to detect global context and local cues by vision transformers in two stages. 

Those learning-based methods can achieve remarkable progress in correctness via integrating features from multi-layers and uncertainty information. However, the generated edge maps are too thick for downstream tasks and heavily rely on the post-processing. Although efforts for crisp edge detection have been made on loss functions~\cite{deng2018learning, huan2021unmixing} and the label refinement strategy~\cite{ye2023delving}, we argue that the community still needs an edge detector that can directly satisfy both correctness and crispness without any post-processing.

\paragraph{Diffusion probabilistic model.} Diffusion models~\cite{sohl2015deep, ho2020denoising, huang2023decoupled} are a class of generative models based on a Markov chain, which gradually recover the data sample via learning the denoising process. Diffusion models demonstrate remarkable performance in fields of computer vision~\cite{nichol2021glide, avrahami2022blended, gu2022vector}, nature language processing~\cite{austin2021structured} and audio generation~\cite{popov2021grad}. Despite those great achievements in generative tasks, diffusion models also have great potential for perception tasks, such as image segmentation~\cite{brempong2022denoising, wu2023medsegdiff} and object detection~\cite{chen2022diffusiondet}.

Inspired by the above pioneers~\cite{xie2015holistically, chen2022diffusiondet, huang2023decoupled}, our method has two main differences to directly generate accurate and crisp edge maps with acceptable inference time. First, we design to impose a learnable Fourier convolution module in the decoupled diffusion architecture, to adaptively filter latent features in Fourier space depending on the target distribution. Second, to keep the pixel-level uncertainty prior from edge datasets with multiple annotators, we distillate the gradients directly to latent space for improved results and stabilized training. The proposed DiffusionEdge, to the best of our knowledge, is the first usage of diffusion models for generic edge detection, and is superior in both correctness and crispness.
\section{Method}

\begin{figure*}
	\centering
	\includegraphics[width=2\columnwidth]{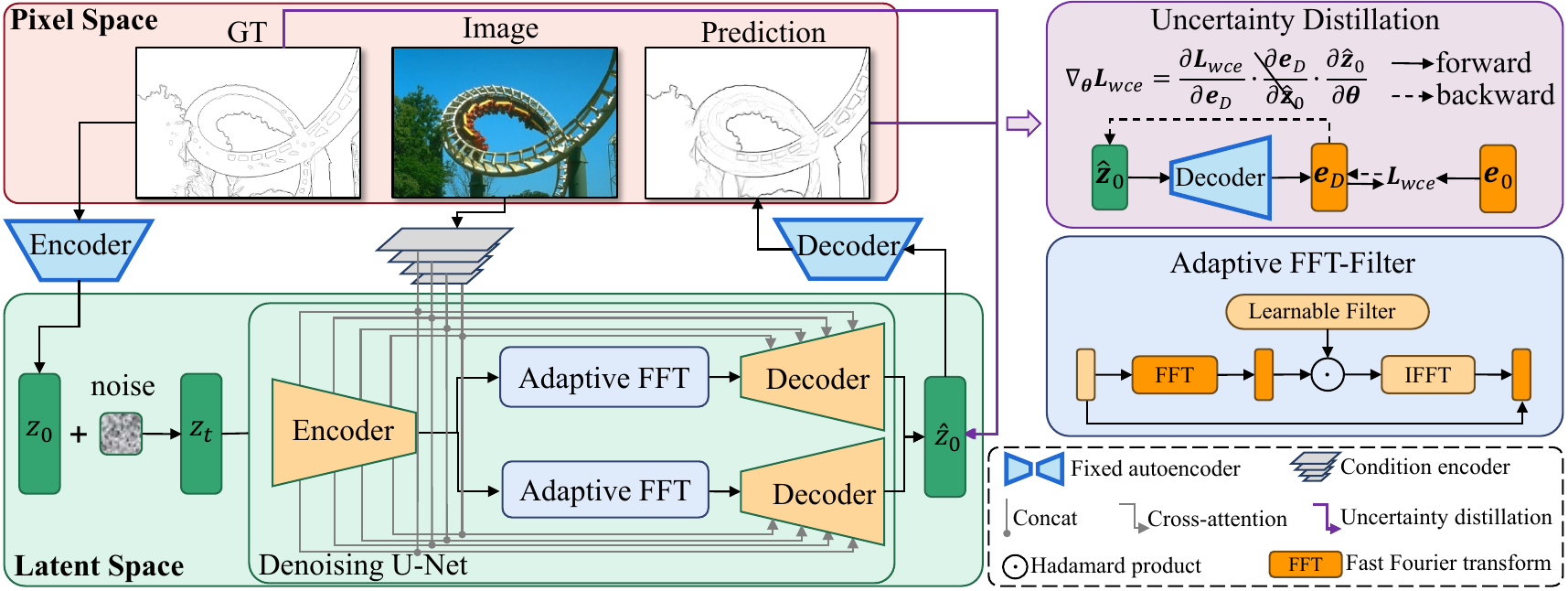}
	\caption{The overall framework of the proposed DiffusionEdge.}
	\label{fig:framework}
\end{figure*}

The overall framework of the proposed DiffusionEdge is illustrated in Figure~\ref{fig:framework}. Inspired by previous works~\cite{rombach2022high, wu2023medsegdiff, huang2023decoupled}, we train the diffusion model with decoupled structure in latent space and take the input image as the extra condition. Based on the diffusion process introduced in preliminaries, we introduce the adaptive FFT-filter for frequency parsing. To keep pixel-level uncertainty from multiple annotators and reduce computational resources, we proposed to directly optimize the latent space with cross-entropy loss in a distillation manner.

\subsection{Preliminaries} 
\label{sec:preliminaries}
Current studies \cite{chen2022diffusiondet, wu2023medsegdiff} have shown the great potential of DPMs in perception tasks, however, it suffers from prolonged sampling time. Inspired by \cite{huang2023decoupled}, we adopt a decoupled diffusion model (DDM) to speed up the sampling process.
The decoupled forward diffusion process is governed by the combination of the explicit transition probability and the standard Wiener process:
\begin{equation}
	q(\mathbf{e}_{t}|\mathbf{e}_{0}) = \mathcal{N}(\mathbf{e}_{0}+\int_{0}^{t} {\mathbf{f}_{t}\mathrm{d}t}, t\mathbf{I}),
	\label{eq1}
\end{equation}
where $\mathbf{e}_{0}$ and $\mathbf{e}_{t}$ are the initial and noisy edges, and $\mathbf{f}_{t}$ is the explicit transition function representing the opposite direction of the gradient of the edge. Following \cite{huang2023decoupled}, we use the constant function as default $\mathbf{f}_{t}$.
The corresponding reversed process is represented by:
\begin{equation}
	\begin{aligned}
		q(\mathbf{e}_{t-\Delta t}|\mathbf{e}_{t}, \mathbf{e}_{0}) &= \mathcal{N}(\mathbf{e}_{t} +\int_{t}^{t-\Delta t} {\mathbf{f}_{t}\mathrm{d}t}\\
		&-\frac{\Delta t}{\sqrt{t}}\boldsymbol{n}, \frac{\Delta t(t-\Delta t)}{t}\mathbf{I}),
	\end{aligned}
	\label{eq2}
\end{equation}
where $\boldsymbol{n}\sim \mathcal{N}(\mathbf{0}, \mathbf{I})$. To train the decoupled diffusion model, we need to supervise the data and noise components simultaneously, therefore, the training objective is parameterized by:
\begin{equation}
	\min\limits_{\boldsymbol{\theta}} \mathbb{E}_{q(\mathbf{e}_{0})} \mathbb{E}_{q(\boldsymbol{n})} [\Vert \mathbf{f}_{\boldsymbol{\theta}}-\mathbf{f}\Vert^{2} + \Vert \boldsymbol{n}_{\boldsymbol{\theta}}-\boldsymbol{n}\Vert^{2}],
	\label{eq3}
\end{equation}
where $\boldsymbol{\theta}$ is the parameter of the denoising network.
Since diffusion models take up too much computational cost in original image space, we follow \cite{rombach2022high} to transfer the training process into latent space with 4$\times$ downsampling spatial size.

As shown in Fig.~\ref{fig:framework}, we first train an autoencoder that consists of an encoder for compressing the edge ground truth to latent code and a decoder for recovering it from the latent code, respectively. Then, in the stage of training denoising U-Net, we fix the weights of the autoencoder and train the denoising process in latent space. The process can be represented as:
\begin{equation}
	\begin{aligned}
		&\mathbf{f}_{\boldsymbol{\theta}}, \boldsymbol{n}_{\boldsymbol{\theta}} = \mathbf{Net}_{\boldsymbol{\theta}}(\mathbf{z}_{t}, t), \\
		&\mathbf{z}_{t} = \mathbf{z}_{0}+\int_{0}^{t} {\mathbf{f}_{t}\mathrm{d}t}+\sqrt{t}\boldsymbol{n},
	\end{aligned}
\end{equation}
where $\mathbf{Net}_{\boldsymbol{\theta}}$ denotes the denoising U-Net, $\mathbf{z}_{0}=\mathcal{E}(\mathbf{e}_{0})$ is the latent code compressed by the encoder of autoencoder, $t$ is the time step. 

We also incorporate several technical designs for edge detection, making it available to obtain accurate and crisp predictions within acceptable inference time.

\subsection{Adaptive FFT-filter} 
\label{sec:adaptive_fft}


The denoising U-Net aims to decouple the noisy input $\mathbf{e}_{t}$ into the denoised data $\mathbf{e}_{0}$ and the noise component $\boldsymbol{n}$. The vanilla convolution layers are adopted as the decoupling operator, to separate the denoised edge maps and noise component from the noisy variable. However, the convolution operators focus more on feature aggregation, and no not adjust the components of specific frequencies. 
Therefore, we introduce a decoupling operator that can filter out different components adaptively.
As shown in the left-top of Figure~\ref{fig:framework}, we integrate the adaptive Fast Fourier Transform filter (Adaptive FFT-filter) into the denoising Unet to filter out edge maps and noise components in the frequency domain. Specifically, given the encoder feature $\mathbf{F}\in \mathbb{R}^{H\times W\times C}$, we first perform 2D FFT along the spatial dimensions, and represent the transformed feature as $\mathbf{F_{c}}=\mathscr{F}[\mathbf{F}], \mathbf{F_{c}}\in\mathbb{C}^{H\times W\times C}$. 
Then, to learn an adaptive spectrum filter, we construct a learnable weight map $\mathbf{W}\in\mathbb{C}^{H\times W\times C}$ and multiply $\mathbf{W}$ to $\mathbf{F_{c}}$. The spectrum filter benefits the training since it can globally adjust the specific frequencies and the learned weights are adaptive for different frequencies of target distributions. With the useless components filtered out adaptively, we project the feature from the frequency domain back to the spatial domain by Inverse Fast Fourier Transform (IFFT). Finally, we adopt a residual connection from $\mathbf{F}$ to avoid filtering useful information out. We can describe the above process by the following equation:
\begin{equation}
	\mathbf{F}_{o} = \mathbf{F} + \mathscr{F}^{-1}[\mathbf{W}\circ\mathbf{F_{c}}],
	\label{eq4}
\end{equation}
where $\mathbf{F}_{o}$ is the output feature, $\circ$ represents the hadamard product.


\subsection{Uncertainty Distillation}
\label{sec:distillation}
Since the numbers of edge and non-edge pixels are highly imbalanced (the majority of pixels are non-edges), HED~\cite{xie2015holistically} propose to apply weighted binary cross-entropy (WCE) loss for optimization, which is further improved by RCF~\cite{liu2017richer} with uncertainty prior from multiple annotators. With $E_{i}$ to be the ground truth edge probability of $i$th pixel, for the $i$th pixel in the $j$th edge map with value $p_{i}^{j}$, the uncertainty-aware WCE loss is calculated as: 
\begin{equation}\label{eq:wce_loss}
	l_{i}^{j}=
	\left\{
	\begin{array}{lll}
		\alpha \cdot \log \left(1-p_{i}^{j}\right), &  if \ E_{i}=0,\\
		0, & if \ 0 < E_{i} < \eta, \\
		\beta \cdot \log E_{i}^{j}, & otherwise,
	\end{array}
	\right.
\end{equation} 
in which
\begin{equation}
	\begin{array}{l}
		\alpha=\lambda \cdot \frac{\left|E^{+}\right|}{\left|E^{+}\right|+\left|E^{-}\right|}, \\
		\beta=\frac{\left|E^{-}\right|}{|E^{+}|+|E^{-}|},
	\end{array}
\end{equation}
where $\eta$ is the threshold to decide uncertain edge pixels in ground truths, and such ambiguous samples will be ignored during subsequent optimization. $E^+$ and $E^-$ denote the number of edge and non-edge pixels in the ground truth edge maps. $\lambda$ is the weight for balancing $E^+$ and $E^-$. The final loss for each edge map is $\mathcal{L}_{wce}=\sum_{i}^{j} l_i^j$.

Ignoring ambiguous pixels during optimization can avoid confusing the network and stabilizing the training process with improved performance. However, it is almost impossible to apply the WCE loss to the latent space with the misalignment in both numerical range and spatial size. In particular, the threshold $\eta$ (generally ranges from 0 to 1) of WCE loss is defined on image space, but the latent code follows the normal distribution and has a various range.  Moreover, the pixel-level uncertainty is hard to be aligned with the encoded and down-sampled latent features of different sizes. 
Therefore, applying the cross-entropy loss directly to latent code inevitably leads to incorrect uncertainty. 

\begin{figure}
	\centering
	\includegraphics[width=1.0\columnwidth]{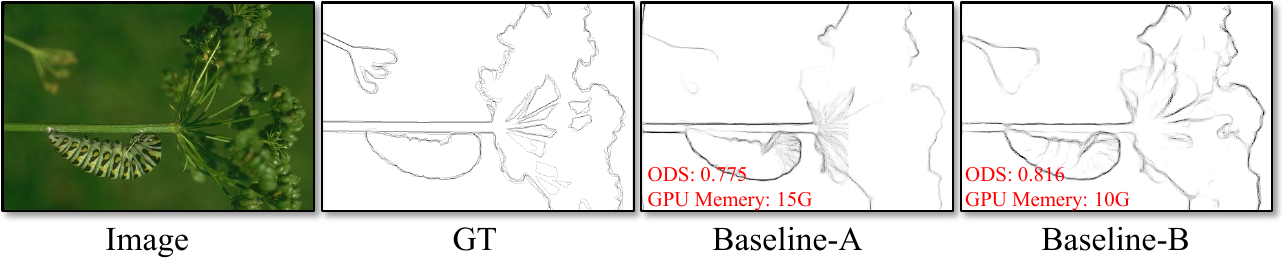}
	\caption{Examples of two baselines with accuracy and memory cost.}
	\label{fig:distillation_toy}
	\vspace{-0.4cm}
\end{figure}

On the other hand, one may choose to decode the latent code back to the image level and thus use the uncertainty-aware cross-entropy to directly supervise the predicted edge maps. Unfortunately, this implementation lets the backward gradient go through the redundant autoencoder, making it hard to feed back effective gradients. Besides, the additional gradient computation in the autoencoder leads to a huge GPU memory cost. 
As shown in Figure~\ref{fig:distillation_toy}, we conduct two experiments to show the negative impact of feeding back the gradient through the autoencoder. We name the setting with gradient through autoencoder Baseline-A. As a comparison, we remove the WCE loss but just use Eq.~\ref{eq3} to supervise the latent code, which is named Baseline-B. The performance of Baseline-B is not satisfactory, and Baseline-A even performs worse with 1.5$\times$ more GPU memory.

To address this problem, we propose the uncertainty distillation loss that can directly optimize the gradient on the latent space.
The results of Baseline-A illustrate that feeding back the gradient through the redundant autoencoder leads to a huge GPU memory cost and hurts the performance, which introduces an inspiration
of eliminating the gradient of autoencoder based on Baseline-B.
Specifically, assuming the reconstructed latent code is $\hat{\boldsymbol{z}}_{0}$, the decoder of the autoencoder is $\mathcal{D}$, and the decoded edge is $\mathbf{e}_{\mathcal{D}}$, we consider the gradient of WCE loss $\mathcal{L}_{wce}$ by the Chain Rule:
\begin{equation}
	\nabla_{\boldsymbol{\theta}}{\mathcal{L}_{wce}} = \frac{\partial\mathcal{L}_{wce}}{\partial\mathbf{e}_{\mathcal{D}}}
	\frac{\partial\mathbf{e}_{\mathcal{D}}}{\partial\hat{\boldsymbol{z}}_{0}}
	\frac{\hat{\boldsymbol{z}}_{0}}{\partial\boldsymbol{\theta}}.
\end{equation}
To remove the negative influence of autoencoder, we skip the gradient through the autoencoder $\partial\mathbf{e}_{\mathcal{D}}/\partial\hat{\boldsymbol{z}}_{0}$ and modify the gradient $\nabla_{\boldsymbol{\theta}}{\mathcal{L}_{wce}}$ by: 
\begin{equation}
	\nabla_{\boldsymbol{\theta}}{\mathcal{L}_{wce}} = \frac{\partial\mathcal{L}_{wce}}{\partial\mathbf{e}_{\mathcal{D}}}
	\frac{\hat{\boldsymbol{z}}_{0}}{\partial\boldsymbol{\theta}}.
\end{equation}

This implementation reduces the computational cost greatly and allows the WCE loss to be applied to latent code directly. In this way, with the time-variant loss weight $\sigma_{t}=(1-t)^{2}$, our final training objective is represented by:
\begin{equation}
	\mathcal{L} = \Vert \mathbf{f}_{\boldsymbol{\theta}}-\mathbf{f}\Vert^{2} + \Vert \boldsymbol{n}_{\boldsymbol{\theta}}-\boldsymbol{n}\Vert^{2} + \sigma_{t}\mathcal{L}_{wce}(\mathbf{e}_{\mathcal{D}}, \mathbf{e}_{0}).
\end{equation}

\section{Experiments}

\subsection{Datasets} 
We conduct experiments on four popular edge detection datasets: BSDS~\cite{arbelaez2010contour}, NYUDv2~\cite{silberman2012indoor}, Multicue~\cite{mely2016systematic} and BIPED~\cite{poma2020dense}.

BSDS consists of 200, 100, and 200 images in the training set, validation set, and test set, respectively. Each image has 4 to 9 annotators and the final edge ground truth is computed by taking their average. 

NYUDv2 is built for indoor scene parsing and is also applied for edge detection evaluation. It contains 1449 densely annotated RGB-D images, and is divided into 381 training, 414 validation and 654 testing images. 

Multicue consists of images from 100 challenging natural scenes. Each image is annotated by several people as well. We randomly split the 100 images into training and evaluation sets, consisting of 80 and 20 images respectively. We repeat the process on Multicue-edge three times and average the scores as the final results.

BIPED contains 250 annotated images of outdoor scenes and is split into a training set of 200 images and a testing set of 50 images. All images are carefully annotated at single-pixel width by experts in the computer vision field.

Previous methods generally augment the dataset with various strategies. For example, images in BSDS are augmented with flipping (2×), scaling (3×), and rotation (16×), leading to a training set that is 96× larger than the original version. Others are concluded in Table~\ref{tab:aug_strategy}. However, our method trains all datasets with only randomly cropped patches of 320$\times$320. In BSDS, we apply random flipping and scaling. In NYUDv2, Multicue and BIPED datasets, only random flipping is adopted.

\begin{table}[htbp]
	\centering
	\scalebox{0.95}{
		\begin{tabular}{c|c}
			\hline
			Datasets &	Augmentation strategies \\
			\hline
			BSDS & F (2×), S (3×), R (16×)=96×\\
			NYUD & F (2×), S (3×), R (4×)=24×\\
			Multicue & F (2×), C (3×), R (16×)=96×\\
			BIPED & F (2×), C (3×), R (16×), G(3×)= 288×\\
			\hline
			
	\end{tabular}}
	\caption{Augmentation strategies adopted on four edge detection benchmarks for previous methods. F: flipping, S: scaling, R: rotation, C: cropping, G: gamma correction.}
	\label{tab:aug_strategy}%
	\vspace{-0.4cm}
\end{table}%

\subsection{Implementation Details} 
We implement our DiffusionEdge using PyTorch~\cite{paszke2019pytorch}.
To train the autoencoder, we collect the edge labels from the training set of all the datasets.
For training the denoising U-Net, we set the smallest time step to 0.0001. We train the models using AdamW optimizer with an attenuated learning rate (from $5e^{-5}$ to $5e^{-6}$) for 25k iterations, and each training takes up about 15 GPU hours. We employ the exponential moving average (EMA) to prevent unstable model performances during the training process.
The balancing weight $\lambda$ and the threshold $\eta$ to identify uncertain edge pixels are set to 1.1 and 0.3, respectively, for all experiments.
We train all datasets with randomly cropped patches of size 320$\times$320 with batch size 16. We conduct inferences with slide 240$\times$240 and take the average value under overlap areas.
All the training is conducted on a single RTX 3090 GPU. 
When inferencing each single image on BSDS dataset, with the sampling Equation~\ref{eq2}, it takes about 3.5GB GPU memory, 1.2 seconds for one-step sampling and 3.2 seconds for five steps on a 3080Ti GPU. 

\subsection{Evaluation Metrics}
To evaluate the precision, recall, and F-score for general edge detection, the predicted edge map should be binarized by an optimal threshold. Following prior works, we compute the F-scores of Optimal Dataset Scale (ODS) and Optimal Image Scale (OIS). ODS employs a fixed threshold throughout the dataset, while OIS chooses an optimal threshold for each image. 
F-scores are computed by $F = \frac{2\cdot P\cdot R}{P+R}$, where $P$ denotes precision and $R$ denotes recall. 
For ODS and OIS, the maximum allowed distances between corresponding pixels from predicted edges and ground truths are set to 0.011 for NYUD and 0.0075 for other datasets. 

To comprehensively evaluate the crispness of edge maps, following previous works~\cite{huan2021unmixing, ye2023delving}, we also report the Standard evaluation protocol (SEval), Crispness-emphasized evaluation protocol (CEval), and the Average Crispness (AC). SEval is calculated after applying a standard post-processing scheme containing an NMS step and a mathematical morphology operation to obtain thinner edge maps. CEval is calculated without any post-processing so that thick edge maps generally get lower precision with more false positive samples. The AC for each edge map is calculated as the ratio of the sum of pixel values after NMS, to the sum of pixel values before NMS, which ranges from 0 to 1. Larger AC means crisper edge maps.

\begin{figure*}[t]
	\centering
	\includegraphics[width=1.9\columnwidth]{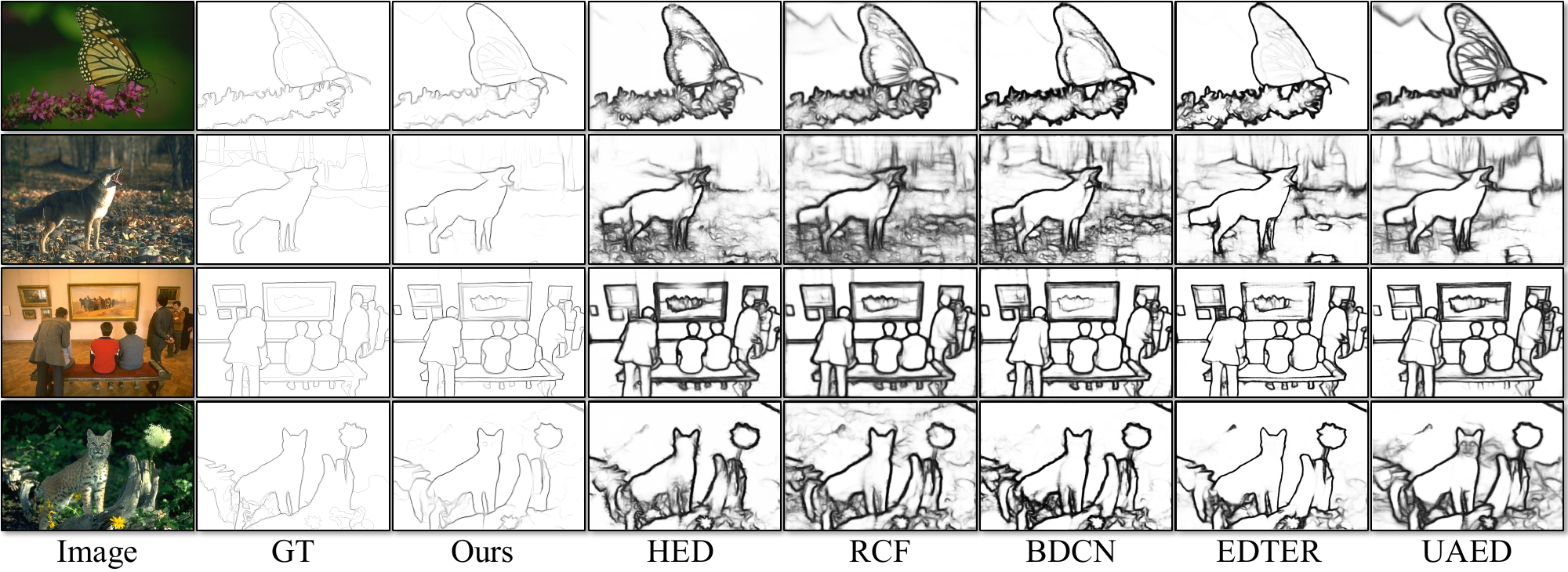} 
	\caption{Qualitative comparisons on BSDS dataset with previous state-of-the-arts. Edge maps generated by our DiffusionEdge are both accurate and crisp with less noise. Zoom-in is highly recommended to observe the details.}
	\label{fig:comparison_bsds}
\end{figure*}

\subsection{Ablation Study} 
\paragraph{The effect of key components.} 
We first conduct experiments to verify the impact of the Adaptive FFT-filter (AF) and Uncertainty Distillation (UD) strategy. The quantitative results are summarized in Table~\ref{tab:ablation_keys}. We can observe that each single AF or UD can promote the performance, while UD is more critical since it plays an important role of optimizing the latent space with valuable uncertainty information. Considering that the AC varies very slightly, the combination of AF and UD achieves the best performance.

\begin{table}[htbp]
	\centering
	\scalebox{0.95}{
		\begin{tabular}{c|c|ccc}
			\hline
			AF &UD & ODS & OIS &AC\\
			\hline
			$\times$&$\times$ &0.816 &0.829 &0.521  \\
			$\times$&$\surd$ &0.831	&0.845 &  0.528\\
			$\surd$&$\times$ &0.825 &0.837 &0.461  \\
			$\surd$&$\surd$ &0.834 &0.848 &0.476  \\
			\hline
			
	\end{tabular}}
	\caption{Ablation study of the effectiveness of the proposed Adaptive FFT-filter (AF) and Uncertainty Distillation (UD) in DiffusionEdge on BSDS dataset. All results are computed with a single scale input, and the same for others. }
	\label{tab:ablation_keys}%
	\vspace{-0.4cm}
\end{table}%

\paragraph{The effect of backbones and diffusion steps.}
We study the impact of different backbones for the image (condition) encoder with ResNet101~\cite{he2016deep}, Effecientnet-b7~\cite{tan2019efficientnet} and Swin-B~\cite{liu2021swin}. Also, the number of iterating steps could be another key parameter in diffusion models. All the results are reported in Table~\ref{tab:ablation_back_step}. We can observe that the crispness varies slightly in all settings, revealing the superiority of DiffusionEdge for crisp edge detection. Swin performs better than other backbones, and we find the number of sampling steps (ranging from 1 to 50) brings litter difference ($<$0.4\% in ODS and OIS) to the final results. Moreover, only one sample step can already achieve state-of-the-art performance. Since more steps mean more inference time, considering all the correctness, crispness and efficiency, we adopt step 5 as the standard setting for all experiments.
\begin{table}[htbp]
	\centering
	\scalebox{0.95}{
		\begin{tabular}{c|ccc}
			\hline
			Backbone & ODS & OIS & AC \\
			\hline
			ResNet &0.823 &0.837 & 0.514 \\
			EffecientNet &0.829	&0.840 & 0.508 \\
			Swin &0.834	&0.848	&0.476\\
			
			\hline
			\hline
			Steps (with Swin) & ODS & OIS & AC \\
			\hline
			Step 1	&0.833	&0.844  &0.453\\
			Step 3	&0.835	&0.847	&0.476\\
			Step 5		&0.834	&0.848	&0.476\\
			Step 10		&0.834	&0.848	&0.476\\
			Step 20		&0.833	&0.847	&0.475\\
			Step 50		&0.833  &0.846	&0.478\\
			\hline
			
	\end{tabular}}
	\caption{The ablations about different backbones and the number of iterating steps for DiffusionEdge.}
	\label{tab:ablation_back_step}%
	\vspace{-0.4cm}
\end{table}%

\subsection{Comparison with State-of-the-arts}

\textbf{On BSDS.}
We compare our model with \textit{traditional detectors} including Canny~\cite{canny1986computational}, SE~\cite{dollar2014fast} and OEF~\cite{hallman2015oriented}, \textit{CNN-based detectors} including N\textsuperscript{4}-Fields~\cite{ganin2014fields}, DeepContour~\cite{shen2015deepcontour}, HFL~\cite{bertasius2015high}, CEDN~\cite{Yang2016Object}, Deep Boundary~\cite{kokkinos2015pushing},
COB~\cite{maninis2017convolutional}, CED~\cite{Wang2018Deep}, AMH-Net~\cite{xu2017learning}, DCD~\cite{liao2017deep}, LPCB~\cite{deng2018learning}, HED~\cite{xie2015holistically}, RCF~\cite{liu2017richer}, BDCN~\cite{he2019bi}, PiDiNet~\cite{su2021pixel}, UAED~\cite{zhou2023treasure} and \textit{the transformer-based detector} EDTER~\cite{pu2022edter}. The best results of all methods are taken from their publications.

\begin{table}[htbp]
	\centering
	\scalebox{0.93}{
		\begin{tabular}{c|cc|cc|c}
			\hline
			\multirow{2}{*}{Methods} & \multicolumn{2}{c|}{SEval} & \multicolumn{2}{c|}{CEval} & \multirow{2}{*}{AC} \\
			\cline{2-5}        & ODS   & OIS   & ODS   & OIS   &  \\
			\hline
			Canny & 0.611 & 0.676 & - & - & -\\
			SE & 0.743 & 0.764 & - & - & -\\
			OEF & 0.746	& 0.770 & - & - & -\\
			N\textsuperscript{4}-Fields & 0.753 & 0.769 & - & - & - \\
			DeepContour & 0.757 & 0.776 & - & - &- \\
			HFL & 0.767 & 0.788 & - & - &- \\
			CEDN &0.788 & 0.804 & - & - &- \\
			DeepBoundary & 0.789 & 0.811 & - & - &-\\
			COB & 0.793 & 0.820 & - & - &-\\
			CED & 0.794 & 0.811 & 0.642 & 0.656 & 0.207\\
			AMH-Net & 0.798 & 0.829 & - & - & -\\
			DCD & 0.799	& 0.817 & - & - &- \\
			LPCB & 0.800 & 0.816 & 0.693 & 0.700 & -\\
			HED & 0.788 & 0.808 & 0.588 & 0.608 & 0.215\\
			RCF & 0.798 &0.815 & 0.585 & 0.604 & 0.189\\
			BDCN & 0.806 & 0.826 & 0.636  & 0.650 & 0.233\\	
			PiDiNet &0.789 & 0.803 & 0.578& 0.587& 0.202\\
			EDTER & 0.824 & 0.841 & 0.698  & 0.706 & 0.288 \\	
			UAED &0.829 &0.847 & 0.722 & 0.731 & 0.227\\
			Ours & \textbf{0.834} & \textbf{0.848} & \textbf{0.749} & \textbf{0.754} & \textbf{0.476} \\
			
			\hline
	\end{tabular}}
	\caption{Quantitative results on the BSDS dataset. 
	For fair comparison, we only list the single-scale results generated by models trained with only BSDS data. Note that other methods are trained with augmented dataset (96×), while we train DiffusionEdge with only random flipping and scaling.}
	\label{tab:exp_bsds}%
\end{table}%

By observing the quantitative and qualitative results in Table~\ref{tab:exp_bsds} and Figure~\ref{fig:comparison_bsds}, several conclusions can be drawn: (a) The proposed method achieves the best results in all settings, especially the AC, which means edge maps generated by DiffusionEdge are much more crisper than other methods; (b) Generally, the performance drop between SEval and CEval is smaller with crisper edge maps (larger AC), it is reasonable that thick edge maps contain many ambiguous false positive edges around true positive ones, evaluating without any post-processing lead to very low precision and thus low F-scores of ODS and OIS; (c) Thanks to the adaptive FFT-filter and uncertainty distillation strategy, our qualitative results perform even better with much less noise and more semantically meaningful contours, especially in challenging scenarios with complicated background and texture.

\begin{figure}
	\centering
	\includegraphics[width=0.95\columnwidth]{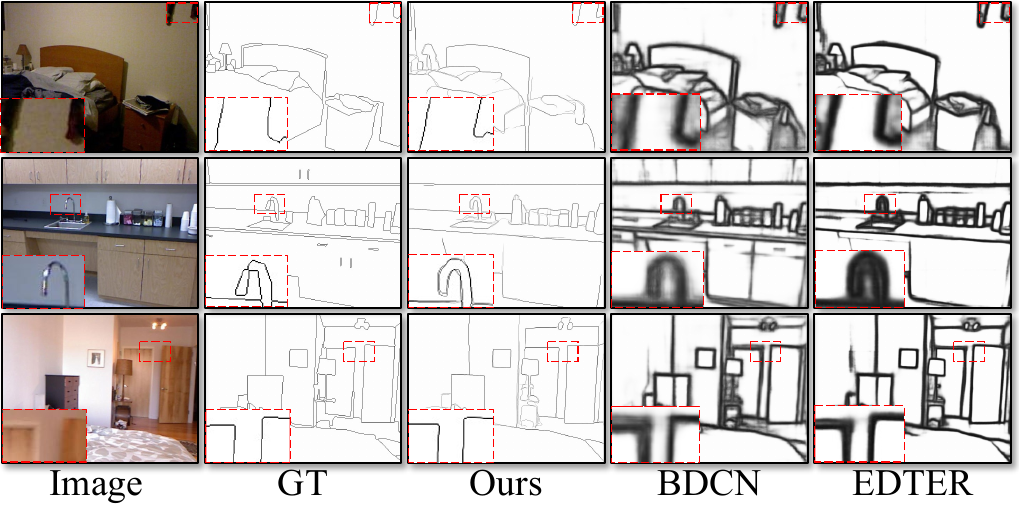} 
	\caption{Qualitative comparisons on NYUDv2 dataset with two state-of-the-art CNN-based and transformer-based methods. Edge maps generated by DiffusionEdge are much crisper and cleaner with competitive performance.}
	\label{fig:comparison_nyud}
	\vspace{-0.4 cm}
\end{figure}

\textbf{On NYUDv2.}
We conduct experiments on RGB images and compare DiffusionEdge with state-of-the-art methods including AMH-Net, LPCB, HED, RCF, BDCN, PiDiNet and EDTER. Quantitative and qualitative results are shown in Table~\ref{tab:exp_nyud} and Figure~\ref{fig:comparison_nyud}, respectively. Our method achieves comparable performance under SEval. However, edge maps generated by other methods are extremely thick with all ACs smaller than 0.2, leading to a significant performance drop under CEval. Such thick edge maps may come from training with the possibly existing label offsets for CNN-based methods~\cite{ye2023delving}. However, DiffusionEdge can directly learn to recover the single-width label and maintain the crispness with slight performance change without post-processing. Consequently, compared to the second best (EDTER), we increase the ODS, OIS of CEval and AC by a large margin of 30.2\%, 28.1\% and 65.1\%, respectively.

\begin{table}[htbp]
	\centering
	\scalebox{0.95}{
		\begin{tabular}{c|cc|cc|c}
			\hline
			\multirow{2}{*}{Methods} & \multicolumn{2}{c|}{SEval} & \multicolumn{2}{c|}{CEval} & \multirow{2}{*}{AC} \\
			\cline{2-5} & ODS   & OIS   & ODS   & OIS   &  \\
			\hline
			
			AMH-Net &0.744 &0.758 & - & - & -\\
			LPCB &0.739 &0.754 &- &- & -\\
			HED & 0.722 &0.737 &0.387 &0.404  & -\\
			RCF &0.745 &0.759 &0.398 &0.413  & -\\
			BDCN & 0.748 &0.762 &0.426 &0.450 & 0.162\\	
			PiDiNet &0.733 &0.747 & 0.399 &0.424 & 0.173\\
			EDTER & \textbf{0.774} &\textbf{0.789} & 0.430  & 0.457 &  0.195\\			
			
			Ours & 0.761 & 0.766 & \textbf{0.732}  & \textbf{0.738} & \textbf{0.846} \\
			
			\hline
	\end{tabular}}
	\caption{Quantitative comparisons on NYUDv2. All results
		are computed with a single scale input. Note that other methods are trained with augmented dataset (24×), while we train DiffusionEdge with only random flipping.}
	\label{tab:exp_nyud}%
\end{table}%

\begin{figure}
	\centering
	\includegraphics[width=1.0\columnwidth]{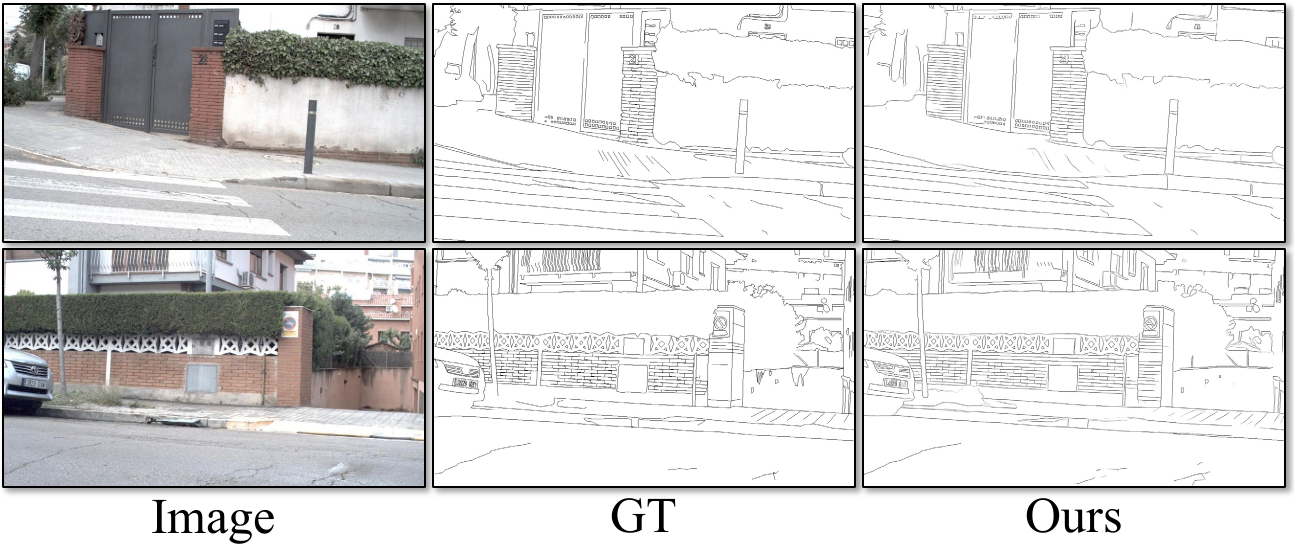} 
	\caption{Qualitative examples on BIPED dataset.}
	\label{fig:comparison_biped}
	\vspace{-0.2cm}
\end{figure}

\textbf{On Multicue and BIPED.}
We further compare DiffusionEdge with HED, RCF, BDCN, DexiNed, PiDiNet, EDTER and UAED, on the datasets of Multicue-edge and BIPED, via the standard evaluation procedure. As shown in Table~\ref{tab:exp_multicue_biped}, our method is superior in both correctness and crispness. It is worth noting that our method achieves a high AC of 0.849 on the BIPED dataset, which means the edges are almost all single-width with no ambiguity, as demonstrated in Figure~\ref{fig:comparison_biped}. Such a success reveals the great potential to directly adopt the predicted results of DiffusionEdge without any post-processing for downstream tasks.

\begin{table}[htbp]
	\centering
	\scalebox{0.93}{
		\begin{tabular}{c|ccc|ccc}
			\hline
			\multirow{2}{*}{Methods} & \multicolumn{3}{c|}{Multicue dataset} & \multicolumn{3}{c}{BIPED dataset} \\
			\cline{2-7} & ODS   & OIS & AC  & ODS   & OIS   & AC  \\
			\hline
			
			Human &0.750 & - & - & -  & - & - \\
			Multicue &0.830 &- & - & -&- &-\\
			HED & 0.851 &0.864 & - & 0.829  &0.847 & -\\
			RCF &0.851 &0.862  & - & 0.843 &0.859 &-\\
			BDCN & 0.891 &0.898 & - &0.839 &0.854 &-\\	
			DexiNed & 0.872& 0.881 &0.274 & 0.859& 0.867&0.295\\
			PiDiNet &0.874 &0.878 & 0.204 & 0.868&0.876&0.232\\
			EDTER & 0.894 &0.900 &  0.196 & 0.893 & 0.898 & 0.26\\			
			UAED &0.895 & 0.902 & 0.211 &- &- &-\\
			Ours &\textbf{0.904} & \textbf{0.909} & \textbf{0.462} &\textbf{0.899} &\textbf{0.901} & \textbf{0.849}\\
			
			\hline
	\end{tabular}}
	\caption{Quantitative comparisons on Multicue and BIPED. All results are computed with a single scale input.}
	\label{tab:exp_multicue_biped}%
\end{table}%

\textbf{On Crispness.}
To further verify the superiority of DiffusionEdge for crisp edge detection, we compare the AC of our method and other strategies proposed for generating crisp edge maps. Here we apply the Dice loss~\cite{deng2018learning} (“-D” in table), the tracing loss~\cite{huan2021unmixing} (“-T” in table) and the Guided Label Refinement~\cite{ye2023delving} (“-R” in table) based on PiDiNet~\cite{su2021pixel}.  As shown in Table~\ref{tab:comparision_crispness}, our DiffusionEdge achieves the best crispness in all cases compared with other methods. Although much efforts have been made for improving the crispness of CNN-based networks (PiDiNet here as an example), the crispness is still limited by the encoder-decoder architecture in nature. However, the diffusion-based edge detection scheme recovers edge maps directly on the original size and the predictions can be almost as crisp as the ground truths.

\begin{table}[htbp]
	\centering
	\begin{tabular}{c|ccc}
		\hline
		\multirow{2}{*}{Methods} & \multicolumn{3}{c}{AC} \\
		\cline{2-4}         & BSDS  & Multicue & BIPED \\
		\hline
		PiDiNet-D & 0.306 & 0.208 & 0.34 \\
		PiDiNet-T & 0.333 & 0.217 & 0.296 \\
		PiDiNet-R & 0.424 & 0.424 & 0.512 \\
		Ours  & \textbf{0.476} & \textbf{0.462} & \textbf{0.849} \\
		\hline
	\end{tabular}
	\caption{Comparisons of the average crispness (AC) on BSDS, Multicue and BIPED dataset with the backbone of PiDiNet. “-D”, “-T” and “-R” means training with dice loss, tracing loss and training with refined labels, respectively.}
	\label{tab:comparision_crispness}
\end{table}%

\section{Conclusions and Limitations}
In this paper, we introduce the first diffusion-based network for crisp edge detection. With several technical designs including the adaptive FFT-filter and uncertainty distillation strategy, our DiffusionEdge is able to directly generate accurate and crisp edge maps without any post-processing. 
Extensive experiments demonstrate the superiority of DiffusionEdge both quantitatively and qualitatively. The crispness is even satisfactory enough and shows the potential for benefiting subsequent tasks in an end-to-end manner.

\paragraph{Limitations.}
The correctness and crispness of edge maps extracted by DiffusionEdge can be simultaneously qualified for downstream tasks. 
However, another one of the three challenges, the efficiency, remains an open problem. Improving the diffusion model for faster inference speed is still a promising future direction to explore.

\section{Acknowledgments}
This work is supported in part by the NSFC (62172155, 62072465, 62325221, 62132021, 62002375, 62002376), the National Key Research and Development Program of China (2018AAA0102200), the Natural Science Foundation of Hunan Province of China(2021RC3071, 2022RC1104, 2021JJ40696) and the NUDT Research Grants (ZK22-52).

\bibliography{aaai24}

\end{document}